# AUTOMATIC COARSE CO-REGISTRATION OF POINT CLOUDS FROM DIVERSE SCAN GEOMETRIES: A TEST OF DETECTORS AND DESCRIPTORS

F. Pirotti[1,2]*, A. Guarnieri[1,2], S. Chiodini[3,4], C. Bettanini[3,4]

[1] Department of Land, Environment and Agro-Forestry (TESAF), University of Padova, Legnaro, Italy
(francesco.pirotti,alberto.guarnieri)@unipd.it
[2] Interdepartmental Research Center of Geomatics (CIRGEO), University of Padova, Legnaro, Italy
[3] Department of Industrial Engineering (DII), University of Padova, via Venezia 1, Padova, Italy
(sebastiano.chiodini,carlo.bettanini)@unipd.it
[4] Center for Studies and Activities for Space "Giuseppe Colombo" (CISAS), University of Padova, via Venezia 15, Padova, Italy

**ABSTRACT:**

Point clouds are collected nowadays from a plethora of sensors, some having higher accuracies and higher costs, some having lower accuracies but also lower costs. Not only there is a large choice for different sensors, but also these can be transported by different platforms, which can provide different scan geometries. In this work we test the extraction of four different keypoint detectors and three feature descriptors. We benchmark performance in terms of calculation time and we assess their performance in terms of accuracy in their ability in coarse automatic co-registration of two clouds that are collected with different sensors, platforms and scan geometries. One, which we define as having the higher accuracy, and thus will be used as reference, was surveyed via a UAV flight with a Riegl MiniVUX-3, the other on a bicycle with a Livox Horizon over a walking path with un-even ground.

The novelty in this work consists in comparing several strategies for fast alignment of point clouds from very different surveying geometries, as the drone has a bird's eye view and the bicycle a ground-based view. An added challenge is related to the lower cost of the bicycle sensor ensemble that, together with the rough terrain, reasonably results in lower accuracy of the survey. The main idea is to use range images to capture a simplified version of the geometry of the surveyed area and then find the best features to match keypoints. Results show that NARF features detected more keypoints and resulted in a faster co-registration procedure in this scenario whereas the accuracy of the co-registration is similar to all the combinations of keypoint detectors and features.

## 1. INTRODUCTION

Point clouds are becoming ubiquitous across different fields of applications (Kutchartt et al., 2022), not only for research but especially for practical applications such as public administrations and end-users which need down-stream usage of 3D point clouds. Lower cost of laser scanners and a much higher portability are causing an increase in the availability of scans which have lower accuracies than higher-end (and more expensive) solutions, but still provide important significant information for visualization and for measurements at under determined scales of representation depending on the respective tolerance of the scale to the measurement error.

When a higher-accuracy scan is available, it can be used as reference for the co-registrations process. The well-known procedure consists in either manual or semi-automatic detection of corresponding tie points between the two scans, aligning one scan on the other according to a roto-translation matrix that is estimated usually by solving multiple least squares fitting after ignoring gross outliers via a RANSAC approach.

One key factor to address when defining a co-registration workflow is the size of the data. It is trivial to say that the advance in technologies has led to denser and larger point clouds. This increases computing requirements, especially considering that keypoints are detected by considering neighbourhoods around each candidate point. Neighbourhoods imply the specification of a minimum-maximum number of nearest neighbours or a sphere radius around which to determine the number of neighbours to take into consideration. This results in each keypoint determining a local point cloud from which certain descriptors are extracted that provide a feature vector at each keypoint. Feature vectors are then used to match keypoints between two sets, one from the reference point cloud, which will not move, and one from the registered (to be) point cloud. Matching methods use different approaches, but all use the feature vector to check for best similarity.

In the following work we exploit state-of-the-art algorithms together with some modifications to allow for fast and automatic coarse co-registration of two point clouds: (1) a more accurate point cloud from a state-of-the-art drone flight and laser scan sensor from Riegl MiniVUX-3 (2) a point cloud from a low-cost setup that was transported by means of a bicycle. In the next sections we report the steps and the solutions adopted for each step, highlighting pros and cons of the choices and providing some benchmarking in terms of calculation time and accuracy resulting from the different combinations. In total four keypoint detectors were tested, SIFT, Harris, ISS and NARF and three feature descriptors Spin Images FPFH and NARF.

---
\* Corresponding author

## 2. MATERIALS AND METHODS

### 2.1 Study area

The test area is the historical garden of Villa Revedin Bolasco located in Castelfranco Veneto (TV) in the Veneto Region of Italy. The point cloud represents the garden area which is approximately 8 ha, with a lake in the middle and several cultural heritage elements such as statues, buildings and trees of different sizes. The vegetation varies from dense evergreen trees to broadleaves. There is a walking path around the lake totalling about 1 km in length. The path was used for the ground scanning from the bicycle. The ground of the walking path is uneven and gravel-based, which causes light mobile platforms, such as the bicycle used for the survey, to vibrate while moving.

Targets were placed around the garden (Figure 1 left) and measured with a total station and a GNSS with RTK resulting in an horizontal and vertical accuracy (one sigma) of 2.2 cm and 2.7 cm respectively see (Pirotti et al. 2022) for more info. The targets where amongst the ones used to check the point cloud from the UAV flight with the MiniVUX-3.

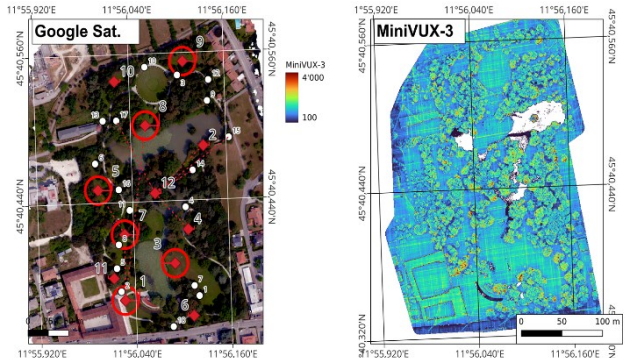

**Figure 1.** Left – overview of the garden and GCP target positions in red and white circles: red circles are big targets; white circles are smaller targets. Right – overall density.

### 2.2 Reference - UAV LiDAR survey

One scan was acquired via Riegl MiniVUX-3 from a drone flight over the area. The UAV carrier consisted in a Soleon LasCo X8 multicopter equipped with one of the three sensors (one camera and two laser scanners) for each flight. Positioning and orientation were measured via a GNSS with RTK corrections and IMU (Applanix APX-20). The LiDAR point cloud was obtained from the Riegl MiniVUX-3UAV sensor that provided a pulse repetition rate of 200,000 measurements per second. This resulted in a density of 1000 point per square meters in average (Figure 1 right). See (Pirotti et al., 2022) for more information on the survey.

This scan will be referred to as "reference" and it will be used as the scene towards which the other cloud, referred to as "registered" is to be aligned to.

### 2.3 Ground survey via bicycle

The second scan was done using a low-cost solution implemented in-house and consisting in an ensemble suite. The sensor suite used for trajectory reconstruction includes the following instruments:

- 3D LiDAR LiVOX Horizon, a newly released solid-state LiDAR, designed for vehicular perception. Due to its non-repetitive scanning pattern, it reaches a more uniform FoV coverage compared to LiDAR scanning solutions with mechanical spinning system.
- LiDAR six-axis build-in IMU, model BMI088, from which the acceleration and angular velocity signals have been acquired. The IMU signals are synchronised with the LiDAR scans.
- HolyBro Neo-M8N GNSS receiver. The GNSS antenna is connected to a Pixhawk® 4 Mini.

| 3D LiDAR | |
|---|---|
| Model | Livox Horizon |
| Maximum Detection Range | 90 m @ reflectivity 10% <br> 130 m @ reflectivity 20% <br> 260 m @ reflectivity 80% |
| Minimum Detection Range | 0.5 m |
| FOV | 81.7° (Horizontal) × 25.1° (Vertical) |
| Distance Noise | $1\sigma$ (@20 m) < 2cm |
| Angular Noise | $1\sigma$ < 0.05° |
| Beam Divergence | 0.03° (Horizontal) × 0.28° (Vertical) |
| Point Rate | 240000 points/s (First Return) |
| Scanning Rate | 10 Hz |
| Power | 12 W |
| Weight | Approx. 1.3 kg |
| IMU | |
| Model | BMI088[1] |
| Digital resolution | Accelerometer (A): 16-bit <br> Gyroscope (G): 16-bit |
| Resolution | (A): 0.09 mg <br> (G): 0.004°/s |
| Measurement range and sensitivity | ± 3 g: 10920 LSB/g <br> (G): ± 125°/s: 262.144 LSB/°/s |
| Zero offset | (A): ± 20 mg <br> (G): ± 1°/s |
| TCO | (A): ± 0.2 mg/K <br> (G): ± 0.015 °/s/K |
| Noise density (typ.) | (A): 175 µg/√Hz <br> (G): 0.014 °/s/√Hz |
| Acquisition Rate | 200 Hz |
| GNSS | |
| Model | HolyBro Pixhawk 4 Neo-M8N GPS |
| Satellite systems | GPS/QZSS; GLONASS; Galileo; BeiDou |
| Max nav update rate | 5 Hz (Glonass/BeiDou) 10 Hz (GPS) |
| Velocity accuracy | 0.05 m/s |
| Heading accuracy | 0.3° |
| Horizontal accuracy | Autonomous 2.5 m <br> SBAS 2 m |

**Table 1.** Characteristics of the sensor ensemble used for the bicycle survey.

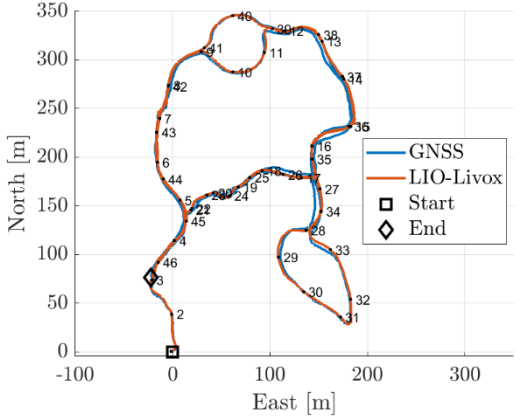

**Figure 2**. Comparison between the LiDAR trajectory reconstructed with the LIO-Livox LiDAR SLAM algorithm and the GNSS reference. The numbers indicate the progression of the route covered.

The mapping method is based on an incremental processing algorithm called LIO-Livox, the method is a LiDAR-SLAM (Simultaneous Localization and Mapping) algorithm which fuses 3D LiDAR scans with the acceleration and angular velocity readings of an IMU (Inertial Measurement Unit) (Li et al., 2021). The algorithm compensates for the motion distortion of the point cloud through IMU readings reintegration. The LiDAR poses are then used to register the captured 3D point clouds together. Subsequently, the LiDAR trajectory is aligned with the East-North-Up (ENU) trajectory measured by a GNSS system. To achieve this alignment, Horn's quaternion-based method (Horn, 1987) is used to estimate the transformation required to bring the reconstructed map into the ENU frame.

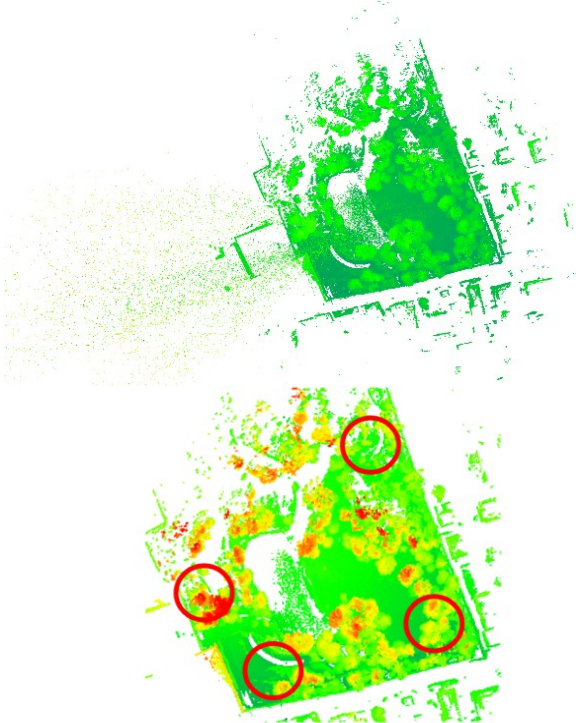

**Figure 2.** Point cloud from the first part of the scan survey with the bicycle before noise removal (top) and after noise removal with the four spheres around check points for accuracy assessment – scaled 10x in the figure (bottom).

Sensors are controlled by a Central Data Management Unit (CDMU), a Raspberry Pi 4 single-board computer, which comprises all electronics for data acquisition and storage. The software is based on ROS middleware, which is a flexible framework used for robotics research. The sensor suite was placed on the front of a bicycle that was used for the acquisitions.

## 3. COARSE CO-REGISTRATION

Coarse alignment can be also seen as the problem of estimating the alignment pose between two models that represent the same object from different perspectives. In our case the model does not change shape, but features are not scanned from the same perspective, and thus are represented differently.

Corresponding points in the two clouds are matched to define an initial coarse co-registration matrix. The point set of the registered point cloud (*Preg*) before and after co-registration is respectively $\{Preg_i^1\}$ and $\{Preg_i^2\}$; i = 1, 2, …, N, where N is the number of points in the registered cloud – in our case the one from the bicycle scan. These two are related by:

$$Preg_i^1 = R \times Preg_i^2 + T + \eta_i \qquad (1)$$

where R is a rotation matrix, T a translation vector, and $\eta_i$ a noise vector.

Tie point matching consists in finding sparse feature correspondences in overlapping areas. It is a key step in photogrammetry, computer vision and remote sensing in general. This can be done manually or by adopting an automatic procedure that identifies keypoints using salient features in multiple scans that have overlapping areas. In this study we use two scans only, but this can be applied to multiple pairs. The two scans used in this study have very different scan geometries and thus provide a complex scenario for this step and also for the next step which consists in calculating descriptive features for each keypoint.

The set of keypoints in the two scans, the reference (*Pref*) and the registered (*Preg*) are $K_i^{Preg}$ and $K_j^{Pref}$ with cardinality N and M respectively i = 1, 2, …, N and j=1, 2, …, M. Keypoints in scan pairs do not necessarily have the same cardinality as many will not match. It must be noted that one keypoint might not have a matching sibling in the other scan, due to incomplete overlaps or simply because of different scan geometries that might provide different structures and thus different descriptive features. This is the case for this study, as the geometry of the drone scan is quite different from the geometry of the bicycle scan. The former is a flying platform and thus catches well the top-most geometries of the area. The latter scans from about one meter from the ground and therefore catches objects from a completely different viewing angle. It must be noted however that the density of the point cloud is such that the drone scan reaches objects under the canopies and the bicycle scan also provides a dense representation of details of the objects in the park.

Corresponding matches between keypoints are found by comparing some kind of distance metric of the corresponding feature vectors. To do this a N x M matrix pairing all keypoints in *Preg* with the all keypoints in *Pref* is used to calculate feature distances for each pair. This can be considered a a graph matching problem that finds correspondences between two sets of features. The graph nodes consist in the keypoints, and each node's attributes are the features extracted from each point.

The methods were implemented using the point cloud library (PCL) developing a specific command-line tool for automatic application of the different methods and extracting accuracy metrics. The C++ code is available upon request.

## 3.1 Simplification of point cloud

Keypoint detectors require to analyse neighbourhoods and are therefore expensive computationally. Our approach to increase speed while keeping the scan complexity was to convert the master and slave point clouds to range images for initial coarse co-registration. Range images can be obtained from point clouds by a simple projection to a XY plane, which can be regarded as the image plane in a pin-hole camera, without lens distortion. The range camera position is defined in our case as being in the center of the point cloud, with a height above the cloud defined to be 5x the maximum length of the bounding box of the cloud. The camera is looking downwards and this roughly corresponds to catching all the points with a field of view (FOV) of 11.4°. If the camera position and orientation is known, calculating the x and y position in the local sensor frame using the collinearity equations is trivial.

$$x - x_0 = -c \frac{r_{11}(X-X_0) + r_{21}(Y-Y_0) + r_{31}(Z-Z_0)}{r_{13}(X-X_0) + r_{23}(Y-Y_0) + r_{33}(Z-Z_0)}$$

$$y - y_0 = -c \frac{r_{12}(X-X_0) + r_{22}(Y-Y_0) + r_{32}(Z-Z_0)}{r_{13}(X-X_0) + r_{23}(Y-Y_0) + r_{33}(Z-Z_0)}$$

Where r is the 3x3 rotation matrix, $X_0\ Y_0\ Z_0$ the projection center and c the distance of the projection center to the plane, or the focal length if the sensor is an actual camera. In our case, the size of the sensor is defined by the angular resolution of the simulated range image, which was set to 0.01° which results in roughly over a million points. Of course, being a range image, not all points are projected to the final range image, but only the ones that are closer to the camera in each of the 0.01° conical rays projected from the camera. The range image is actually a point cloud with less points with respect to the original one, where occluded points are not present in the final point cloud.

## 3.2 Keypoint detectors

Keypoints are locations in the point cloud that represent salient features in a patch around a defined radius. The most important parameter for determining the definition of a keypoint in the point cloud is the support size (Ss). This is the diameter of the sphere used for the calculation of the descriptors that determine if the point can be labelled as a keypoint, depending on their "saliency" which is determined by the algorithm itself. In this work we tested Harris corner detector (HARRIS) Intrinsic Shape Signature (ISS), Scale Invariant Feature Transform (SIFT) and Normal Aligned Radial Feature (NARF).

### 3.2.1 Harris corner detector (HARRIS)
Detecting geometric features in images has been around for a while. Using image gradients by applying an 8-way local maximum to an image transformed by and edge detection filter (e.g. Canny filter) and then applying thresholds to remove edges that are not corners (Harris and Stephens, 1988). What is done using image gradients can be replicated using surface normals in a point cloud thus using the 3D information intrinsic to the data.

### 3.2.2 Intrinsic Shape Signature (ISS)
Intrinsic Shape Signature (ISS) computes the Eigenvalue decomposition of the scatter-matrix of the points within the supporting patch in order to highlight local geometries exhibiting a prominent principal direction (Zhong, 2009).

### 3.2.3 Scale Invariant Feature Transform (SIFT)
Without doubt the most well-known keypoint detector is SIFT, the Scale Invariant Feature Transform detector by Lowe (2004). SIFT takes several parameters, such as (i) number of octaves, (ii) number of scales per octave, (iii) minimum scale and (iv) minimum contrast. Most parameters are independent from target scenes, the only exception being the min scale parameter which is meant to match the point cloud resolution (Teng et al., 2022).

### 3.2.4 Normal Aligned Radial Feature (NARF)
This feature detector was designed for range images. It uses object boundary information by extracting features in areas where the surface has substantial changes in the local neighbourhood (Steder et al., 2010). In this study it was tested to see if the application of NARF to the simulated range image can provide significant results in terms of speed and accuracy of the co-registration. It must be noted that the scene that was scanned, i.e. the area with the garden and buildings, only provide a few features with well-defined boundaries. There are only a few buildings common to the reference and registered scenes and sometimes the margins of the buildings are occluded by tree branches. Tree canopy margins can also play a potential role in defining keypoints by acting as natural boundaries, but these tend to be irregular and thus might be challenging to detect.

## 3.3 Feature descriptors

Feature descriptors should provide information around a keypoint that can significantly describe the area around it in a way that make it unique to that position. This can be seen as a signature that can be compared to find matching points via similarity metrics. Due to different poses of the same scanned objects, these features are ideally invariant to scale and affine transforms. There exist many 3D feature descriptors that provide a feature vector of unique values for a point.

The feature vector is commonly converted to a histogram of metrics that depend on the specific method used. Reducing to a histogram is performed to reduce the descriptor size by binning values into a histogram. The number of values in the vector defines the cardinality of the descriptor. In this study we compared three different feature descriptors: (i) FPFH, (ii) SPIN (iii) NARF, described in better detail below.

### 3.3.1 Fast Point Feature Histograms (FPFH)
These descriptors are based on a histogram that represents neighbours around the keypoint or any point. They are faster than the classical definition of point feature histograms because they have a looser interpretation of neighbours, as they do not consider all of the neighbours around the radius, but can also consider some points outside the radius. The histogram is a binned description of the local geometry around a point in a 3D point cloud datasets (Rusu et al., 2009). Each feature dimension will use 11 bins for the histogram, for a total of 33 values that describe the geometry around that point.

### 3.3.2 Spin Images (SPIN)
The spin image approach is a 3D shape-based object recognition system for recognition of multiple objects in scenes containing clutter and occlusion (Johnson and Hebert, 1999, 1998). A spin image describes the scene around a point by a histogram of point locations aggregated by sum along cells of an image. An image is basically used as an indexed 2D accumulator. The coordinates are computed for each point in the point cloud that is inside the radius of the spin image. The cell, indexed by row and column in the image, is then incremented. Bilinear interpolation is used to smooth the contribution of each point. This procedure is repeated for all points within the range of the spin image. The resulting image can be thought of as an accumulator. Areas in the image that correspond to cells that accumulated many projected points will have larger values. As long as the size of the cells in the image is greater than the median distance between points in the point cloud, the position of individual points will be averaged out during spin image generation.

### 3.3.3 Normal Aligned Radial Feature (NARF)
NARF is not only a keypoint detector but can also define a vector of feature descriptors around a keypoint which is calculated as follows (from Steder et al., 2010):

- calculate a normal aligned range value patch in the point, which is a small range image with the observer looking at the point along the normal,
- overlay a star pattern onto this patch, where each beam corresponds to a value in the final descriptor, that captures how much the pixels under the beam change,
- extract a unique orientation from the descriptor,
- and shift the descriptor according to this value to make it invariant to the rotation.

The unique orientation together with the normal then defines a local coordinate frame at the point that can be used to collect distances to closest points around 36 directions.

### 3.4 Matching: RANSAC

RANSAC is a well-known method for defining models in the presence of a large number of outliers (Fischler and Bolles, 1981). The pose alignment problem in this case is an affine transformation that preserves scale, in other words a rotation and a translation. We use a pre-rejective iterative procedure implemented in the point cloud library that removes points using a local pose-invariant geometric constraints (Buch et al., 2013; Shan et al., 2004). The pre-rejection performance is tuned by two parameters, the similarity threshold and the inlier threshold. The former takes values between 0 and 1, where one means that a perfect match is required, and 0 means that all points are considered value. The metric considered refers a measure of pose-invariant geometric consistencies of the inter-distances between sampled points on the object and the scene. The inlier threshold is the Euclidean maximum distance for a transformed object point to be defined as correctly aligned to the nearest scene point or not. We se the two parameters to 0.55 and 1.0 m respectively.

Once the keypoints are matched and converted to all practical means to tie points, a least squares solution finds the best fit of the rotation (R) and translation (T) matrices in equation 1. An algorithm for finding the least-squares solution of R and T is commonly based on the singular value decomposition (SVD) of a $3 \times 3$ matrix.

A lost note on a parameter required for both keypoint detection and extraction of feature descriptors, i.e. the size of the neighbourhood around which to extract features for determining saliency for determining a candidate keypoint and for the feature vector. We used a 2.0 meter radius for all calculations.

## 4. RESULTS

The first part of the results consists in listing the number of keypoints that were detected and the alignment time, if successful.

| Keypoints | N. keypoints | | Feature Descriptor Alignment time (s) | | |
|---|---|---|---|---|---|
| | Pref | Reg | FPFH | SPIN | NARF |
| ISS | 21943 | 3753 | 18 | F | 11 |
| HARRIS | 1069 | 513 | F | F | F |
| SIFT | 2001 | 1023 | F | F | 12 |
| NARF | 7465 | 2049 | 88 | 12 | 9 |

**Table 2.** Results of number of keypoints and time of calculation of the alignment procedure: *Pref*= reference point cloud (will not move) and *Preg*=registered, (will be aligned). F=failed alignment.

To assess the multiple combinations of keypoint detectors and feature descriptors a visual interpretation would be time consuming thus we defined four spheres around four central points in positions were regular features can provide a persistent reference object. A local cloud-to-cloud distance is then applied. This approach allows to map the estimated differences, that in this case we assume to be errors if we consider the drone flight to be the control cloud. The four points are located near the four furthest points of the area (Figure 2 bottom).

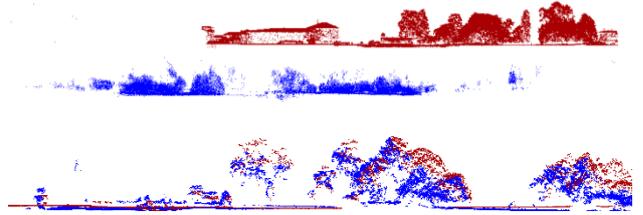

**Figure 3.** Side view before (top) and after successful alignment (bottom).

Figure 3 shows a visual assessment of the alignment from a side view which shows that the alignment is coarse, in the sense that it can be improved by a further closest-point iteration.

| | Alignment accuracy (m) – mean(SD) | | |
|---|---|---|---|
| | **FPFH** | **SPIN** | **NARF** |
| **ISS** | 0.29(0.73) 0.48(0.57) 0.69(0.73) 0.28(0.64) | - | 0.19(0.73) 0.38(0.39) 0.89(0.82) 0.08(0.58) |
| **HARRIS** | - | - | - |
| **SIFT** | - | - | 0.29(0.83) 0.92(0.59) 1.19(0.52) 1.09(0.81) |
| **NARF** | 0.22(0.47) 0.78(0.87) 0.99(1.06) 1.35(0.83) | 0.98(1.23) 0.56(0.87) 0.59(0.81) 1.18(1.74) | 0.79(0.69) 0.18(0.51) 0.29(0.43) 0.18(0.64) |

**Table 2.** Results in terms of mean and standard deviation (SD) of distances between the local point clouds in the four spheres around the check points from *Pref* and *Preg*. The four values start from the bottom right point (see Figure 2 bottom) and go in a clockwise direction

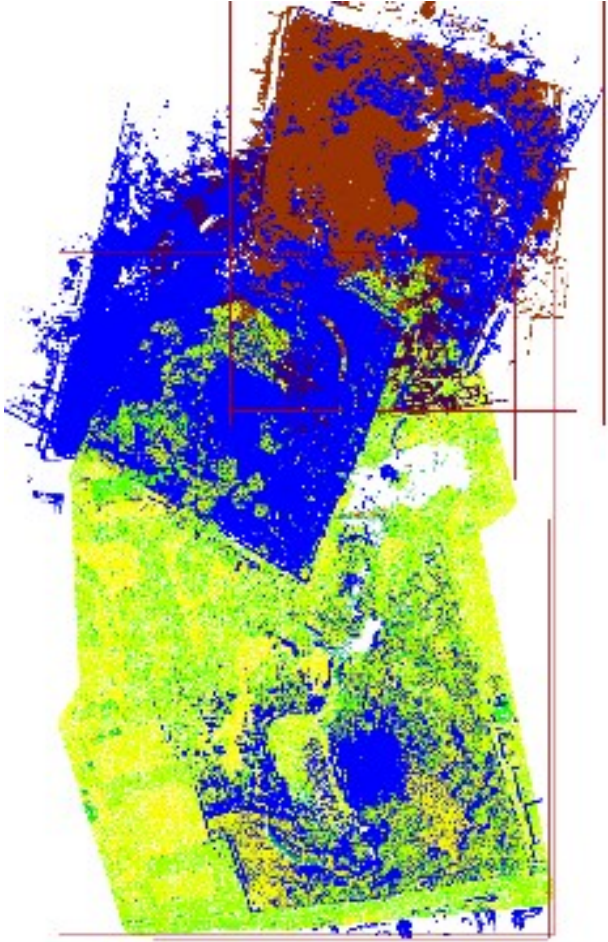

**Figure 4.** Example of wrong (top blue and red point clouds) and correct (bottom blue) alignments.

## 5. DISCUSSION

Results show that in some cases the coarse registration provides a successful RANSAC procedure with very wrong results (Figure. 4). This error is detected by checking the accuracy of the four point clouds in the local spheres, as it results in no matching nearest points between reference and registered cloud, so not accuracy measure. In table 2 and table 3 this is defined as a failed alignment. This is likely a mistake in the tuning of the RANSAC parameters, leading to a loose criterion used to assess a correct RANSAC alignment, in particular the similarity threshold was probably not restrictive enough. Wrong alignments were recorded when using a value of 0.55, thus allowing for a maximal edge length dissimilarity of 45%.

Similar work was done by Hänsch et al., (2014) and it can be noted that in that investigation and in our presented study, the discussion highlights that the number of parameters that can be tuned for each detector and descriptor are numerous, leading to a large amount of combinations that might be tested. In this case we kept a 2.0 m radius for calculating features both for determining the saliency in points for defining keypoints and for feature descriptors, but smaller values might provide a faster procedure with comparable results. Parameter tuning not only changes the overall result, but also changes the initial number of keypoints that are detected. Too many keypoints lead to a very long alignment process and not necessarily to better results. This work set a pipeline that can be used to iterate over a grid of different parameter combinations. It will be the topic of further investigation which will follow this work.

A short note is that for SIFT to calculate on the PCL library, an intensity scalar should be present. Because the two surveys have such a different approach (birds eye view for the UAV survey and ground-view for the bicycle), the real intensity values of objects are not comparable. We therefore assigned values of sphericity to the intensity scalar to be used by SIFT for extracting the keypoints. Sphericity values range from 0 to 1 and this allowed to normalize values across clouds and to set the minimum contrast parameter to 0.2 for detecting a significant number of keypoints.

There are many more keypoint detectors and feature descriptors that are found in literature and implemented in software. The ones presented in this study were chosen as they appeared to have different characteristics worth testing, with other having very similar approaches. It might be argued that this is a subjective choice, and rightly so. Further testing will include more detectors and descriptors, building from the processing pipeline that was implemented in this work.

The NARF detector (Steder et al., 2010) performed slightly better, probably taking advantage of different depths of objects in the range image that was simulated from the original clouds. In NARF object borders, i.e. view-dependent noncontinuous transitions from the foreground to the background, provide robust descriptors that were an advantage in this scenario. Nevertheless, it must be noted that this is not a definite conclusion, as better parameter tuning might provide better accuracy and faster processing time.

The importance of a co-registered point cloud is to be found also in the downstream applications for the point cloud. On this matter the area has been used for several tests regarding the prediction of lighting conditions and relative temperature distribution in the park, where the 3d point cloud model played a central role in defining, via raytracing algorithms, how much solar radiation hits a certain point in space (Pirotti et al., 2022).

## 6. CONCLUSION

In this work we tested 12 combinations of keypoint detectors (4) and feature descriptors (3). All have been tested using a 2 m radius sphere as determinant of the points that are considered for calculating the feature vector that is used for keypoint definition and for the final descriptor element used for matching similar points in the alignment process. We see that NARF both for keypoint detection and as feature descriptor has slightly better performance overall, but we conclude that further investigation regarding better tuning of parameters is necessary for a more rigorous understanding of which approach is better for this scenario. The pipeline proposed in this study will be used in the future for this goal.

Other future work will consist in implementing a fine-registration step with localized keypoint and feature calculations over small subsets around the initial keypoints used in the coarse registration.

## ACKNOWLEDGEMENTS

This work was partly funded by VARCITIES project Grant Agreement number: 869505 —VARCITIES—H2020-SC5-2018-2019-2020/H2020-SC5-2019-2.

## REFERENCES

Buch, A.G., Kraft, D., Kamarainen, J.-K., Petersen, H.G., Krüger, N., 2013. Pose Estimation using Local Structure-Specific Shape and Appearance Context, in: 2013 IEEE International Conference on Robotics and Automation. pp. 2080–2087. https://doi.org/10.1109/ICRA.2013.6630856

Fischler, M.A., Bolles, R.C., 1981. Random Sample Consensus: A Paradigm for Model Fitting with Applications to Image Analysis and Automated Cartography. Commun. ACM 24, 381–395. https://doi.org/10.1145/358669.358692

Hänsch, R., Weber, T., Hellwich, O., 2014. Comparison of 3D interest point detectors and descriptors for point cloud fusion. ISPRS Annals of the Photogrammetry, Remote Sensing and Spatial Information Sciences II–3, 57–64. https://doi.org/10.5194/isprsannals-II-3-57-2014

Harris, C., Stephens, M., 1988. A Combined Corner and Edge Detector, in: Procedings of the Alvey Vision Conference 1988. Presented at the Alvey Vision Conference 1988, Alvey Vision Club, Manchester, p. 23.1-23.6. https://doi.org/10.5244/C.2.23

Horn, B.K.P., 1987. Closed-form solution of absolute orientation using unit quaternions. J. Opt. Soc. Am. A 4, 629. https://doi.org/10.1364/JOSAA.4.000629

Johnson, A.E., Hebert, M., 1999. Using spin images for efficient object recognition in cluttered 3D scenes. IEEE Transactions on Pattern Analysis and Machine Intelligence 21, 433–449. https://doi.org/10.1109/34.765655

Johnson, A.E., Hebert, M., 1998. Surface matching for object recognition in complex three-dimensional scenes. Image and Vision Computing 16, 635–651. https://doi.org/10.1016/S0262-8856(98)00074-2

Kutchartt, E., Pedron, M., Pirotti, F., 2022. ASSESSMENT OF CANOPY AND GROUND HEIGHT ACCURACY FROM GEDI LIDAR OVER STEEP MOUNTAIN AREAS. ISPRS Annals of the Photogrammetry, Remote Sensing and Spatial Information Sciences V-3–2022, 431–438. https://doi.org/10.5194/isprs-annals-V-3-2022-431-2022

Li, K., Li, M., Hanebeck, U.D., 2021. Towards High-Performance Solid-State-LiDAR-Inertial Odometry and Mapping. IEEE Robotics and Automation Letters 6, 5167–5174. https://doi.org/10.1109/LRA.2021.3070251

Lowe, D.G., 2004. Distinctive Image Features from Scale-Invariant Keypoints. International Journal of Computer Vision 60, 91–110. https://doi.org/10.1023/B:VISI.0000029664.99615.94

Pirotti, Francesco, Piragnolo, M., D'Agostini, M., Cavalli, R., 2022. Information Technologies for Real-Time Mapping of Human Well-Being Indicators in an Urban Historical Garden. Future Internet 14, 280. https://doi.org/10.3390/fi14100280

Pirotti, F., Piragnolo, M., Vettore, A., Guarnieri, A., 2022. COMPARING ACCURACY OF ULTRA-DENSE LASER SCANNER AND PHOTOGRAMMETRY POINT CLOUDS. Int. Arch. Photogramm. Remote Sens. Spatial Inf. Sci. XLIII-B1-2022, 353–359. https://doi.org/10.5194/isprs-archives-XLIII-B1-2022-353-2022

Rusu, R.B., Blodow, N., Beetz, M., 2009. Fast Point Feature Histograms (FPFH) for 3D registration, in: 2009 IEEE International Conference on Robotics and Automation. pp. 3212–3217. https://doi.org/10.1109/ROBOT.2009.5152473

Shan, Y., Matei, B., Sawhney, H.S., Kumar, R., Huber, D., Hebert, M., 2004. Linear model hashing and batch RANSAC for rapid and accurate object recognition, in: Proceedings of the 2004 IEEE Computer Society Conference on Computer Vision and Pattern Recognition, 2004. CVPR 2004. p. II–II. https://doi.org/10.1109/CVPR.2004.1315153

Steder, B., Rusu, R.B., Konolige, K., Burgard, W., 2010. NARF: 3D Range Image Features for Object Recognition. Presented at the IEEE/RSJ Int. Conf. on Intelligent Robots and Systems (IROS);

Teng, H., Chatziparaschis, D., Kan, X., Roy-Chowdhury, A.K., Karydis, K., 2022. Centroid Distance Keypoint Detector for Colored Point Clouds. https://doi.org/10.48550/arXiv.2210.01298

Zhong, Y., 2009. Intrinsic shape signatures: A shape descriptor for 3D object recognition, in: 2009 IEEE 12th International Conference on Computer Vision Workshops, ICCV Workshops. Presented at the 2009 IEEE 12th International Conference on Computer Vision Workshops, ICCV Workshops, pp. 689–696. https://doi.org/10.1109/ICCVW.2009.5457637